\pdfoutput=1
\documentclass[sigconf, nonacm]{acmart}

\usepackage{booktabs} 
\usepackage{hyperref}

\usepackage{graphicx} 
\usepackage{amssymb} 
\usepackage{amsmath} 
\usepackage{commath} 
\usepackage{mathtools} 
\usepackage{xcolor} 
\usepackage{cleveref} 
\usepackage{latexsym} 
\usepackage{numprint} 

\usepackage{array} 
\usepackage{multirow} 
\usepackage{xspace} 
\usepackage{stmaryrd} 

\newcommand{\nlm}[0]{${\mathcal NL}$\xspace}

\newcommand{\prem}[1]{{\scriptsize {PD:  {\color{blue} #1}}}}

\newcommand{\etal}{{\em et al.}\xspace}

\newcounter{RQCounter}
\newcommand{\RQ}[2]{%
\refstepcounter{RQCounter} \label{#1}
    \noindent \textbf{RQ\arabic{RQCounter}.}~{\em #2  }                                                                                                        
}

\begin{document}
\title{Do People Prefer ``Natural'' code?}

\author{Casey Casalnuovo}
\affiliation{%
  \institution{Department of Computer Science,\\ UC Davis}
 \streetaddress{2063 Kemper Hall, One Shields Avenue}
  \postcode{95616-8562}
}
\email{ccasal@ucdavis.edu}

\author{Kevin Lee}
\affiliation{%
  \institution{Department of Computer Science,\\ UC Davis}
 \streetaddress{2063 Kemper Hall, One Shields Avenue}
  \postcode{95616-8562}
}
\email{kllee@ucdavis.edu}

\author{Hulin Wang}
\affiliation{%
  \institution{Department of Computer Science,\\ UC Davis}
 \streetaddress{2063 Kemper Hall, One Shields Avenue}
  \postcode{95616-8562}
}
\email{hulwang@ucdavis.edu}

\author{ Prem Devanbu}
\affiliation{%
  \institution{Department of Computer Science,\\ UC Davis}
 \streetaddress{2063 Kemper Hall, One Shields Avenue}
  \postcode{95616-8562}
}
\email{ptdevanbu@ucdavis.edu}

\author{Emily Morgan}
\affiliation{%
  \institution{Department of Linguistics,\\ UC Davis}
 \streetaddress{469, Kerr Hall, One Shields Avenue}
  \postcode{95616-8562}
}
\email{eimorgan@ucdavis.edu}

\begin{abstract}
Natural code is known to be very repetitive (much more so than natural language corpora); 
furthermore, this repetitiveness persists, even after accounting for the simpler syntax
of code. However, programming languages are \emph{very} expressive,
allowing a great many different ways (all clear and unambiguous) to express even very simple computations. 
So why is natural code repetitive? We hypothesize that the reasons for this lie
in fact that code is \emph{bimodal}: it is executed by machines, \emph{but also 
read by humans}. This bimodality, we argue, leads developers
to write code in certain  preferred ways that would be familiar to code readers. 
To test this theory, we {\bf \emph{1)}} model \emph{familiarity} using a language model estimated
over a large training corpus and {\bf \emph{2)}} run an experiment applying several \emph{meaning preserving transformations} 
 to Java and Python expressions in a distinct test corpus to see if forms more familiar to readers (as predicted by the language models) are in fact the ones actually written. 
We find that these transformations generally produce program structures that are less common in practice,
supporting the theory that the high repetitiveness in code is a matter of deliberate preference. 
Finally, {\bf \emph{3)}} we use a human subject study to show alignment between language model score and human preference for the first time in code,
providing support for using this measure to improve code.
 
\end{abstract}

%
%

\maketitle
\section{Introduction and Background}
\label{sec:intro}

Programming languages are highly flexible, and provide many different ways to express even the simplest computation. 
However, despite this flexibility, in practice code is highly repetitive, 
far more so than natural human language. 
This property of ``naturalness''~\cite{Hindle2012} has 
led to the successful application of 
probabilistic methods from NLP and machine learning to problems in software engineering, including
 code completion~\cite{Tu2014, hellendoorn2017deep,Nguyen2015Graph}, finding defects~\cite{Ray2016}, and
recovering variable names~\cite{Raychev2015, vasilescu2017recovering}. 

Here is a puzzle:  \emph{why is code so repetitive}? Ruling out the obvious: it's not just syntax.
There is strong evidence~\cite{casalnuovo2018studying}
that the simpler syntactic and lexical properties of programming language structure, \emph{per se} are not sufficient to explain this. 
Indeed, Casalnuovo \etal~\cite{casalnuovo2018studying} report that when the markers of syntax are elided similarly
from Code and English, Code becomes relatively \emph{even more} repetitive than English. They also argue that this might arise from conscious choice, citing suggestive evidence of similar, strong, repetitive
structure in other corpora that (like code) might entail more effort to read \& write, such as  legal documents, technical manuals, 
and English as a second language. 

However, Casalnuovo \etal~\cite{casalnuovo2018studying}
didn't control for meaning. Given a particular
meaning, are there different ways to express it? Are some forms preferred? 
For instance, in English, people prefer to say \emph{bread and butter} rather than \emph{butter and bread}~\cite{morgan2016abstract}.  Likewise, one can code a simple increment operation different ways:

 \begin{centering}
  {\small\tt i = i+1;}\hspace{0.5in} {\emph{(or)}}\hspace{0.5in}  {\small\tt i = 1+i;} \\
  \end{centering}

The two forms above are trivially equivalent to a machine; if developers worked
like machines, they should be indifferent to the form.
However, most people would strongly prefer the first! This is noteworthy. 
Unlike natural languages (where the semantics are slippery, and depend in subtle 
ways on form)  programming languages have precisely defined semantics; this means
that even the  simplest computations can be coded many different, but  \underline{entirely equivalent}
ways! Programming languages thus actually provide
programmers with {\em great choice} in forms of expression. So, the plot thickens!! \emph{Given
the many expressive choices available, why is code so repetitive?} We believe this is because
coders have predictable preferences: 
\noindent\fbox{%
    \parbox{3in}{%
   \emph{Even given many forms for coding a computation, developers
still prefer it to be coded in more familiar forms.}
    }%
}


We hypothesize that this preference for more familiar forms
is manifest in large
corpora, and is evidenced by human  code readers. This preference, we believe,  
accounts for why code is so repetitive, despite the affordance provided by programming languages \emph{per se}
to code the same computation in many different ways. 

This paper investigates this hypothesis by triangulating
an observational study with a human-subject experiment. We model
``familiarity" using the probability of occurrence in a large code corpus (the idea
being that developers are more familiar with code they see more frequently), and check
whether this measure can correctly select preferred forms in unseen code; 
we also use a controlled experiment to see if this measure can predict human preference. 

We make the following contributions: 
 \begin{itemize}
 \item
Using an adaptation of the UCL-Edinburgh bimodal model~\cite{barrDualTalk, allamanis2017survey}, we provide a theoretical framing of Naturalness as a preferential choice, where 
 developers deliberately choose to express code in familiar forms, modeled via language model.
  \item
We use transformations to generate alternate meaning equivalent forms of code in a \underline{\emph{held-out}} 
test corpus.
 \item
We evaluate the alternative forms using a language model, estimated on a \emph{training} corpus, and find evidence 
that more familiar (less ``surprising") forms are preferentially deployed in the held-out \emph{test} corpus. 
\item
Using Mechanical Turk, we find that human subjects do prefer forms of code (from the test corpus) that the 
language model indicates are more prevalent in the training corpus.
 \end{itemize}

 The contributions of this paper are primarily of a scientific (rather than Engineering) nature. However, we believe our work
provides a new theoretical framework to contemplate a phenomenon of deep current interest to software
engineering researchers, as well as a novel experimental approach, and strong connections to well-established theories of human cognition
of language.  Furthermore, to our knowledge, this is the first study to demonstrate that language model 
probability, used to guide many code applications,
correlates with human preferences for code in a controlled experiment with human subjects.

\section{Background \& Theory}

Natural language (\nlm) corpora exhibit a high-level of repetitiveness, capturable 
\emph{via} statistical modeling, which has been leveraged
in modern tools for translation, speech recognition, text summarization, \emph{etc}. 
More recently~\cite{Hindle2012} it has been noted that code is also quite repetitive and predictable, and many applications
have been developed (See survey by Allamanis \etal~\cite{allamanis2017survey}).

Textual predictability (of code or \nlm) can be captured using \emph{language models} (LM), which
assign probabilities to a textual utterances (an \emph{utterance} is a usable language fragment,  
\emph{e.g.}, a word in English, or a token in code). 
Probabilities are typically assigned
contextually, \emph{e.g.,} probability of $U$ occurring in context $C$:  $p(U|C)$. 
The more likely an utterance $U$ is in context $C$, the higher the probability. LMs
(\emph{e.g.,} $n$-gram models, PCFGs, RNNs) are trained on a large corpus of text, and then are evaluated
by scoring against a test corpus. Their performance is measured using  a normalized \emph{cross entropy} score (in units of bits), 
which is estimated by taking the 
average \emph{surprisal}\footnote{As a reminder, \emph{surprisal} of an event $\epsilon \in$ an event space $E$, with probability $p(\epsilon)$, 
 is $-log(p(\epsilon))$; entropy is the expectation of surprisal over all $x \in E$.}
  of the utterances $U$ the test corpus $T$: 
\[ \frac{1}{\mid T \mid} \sum_{U \in T} -log_2( p (U))~~  \]
where surprisal is the negative log probability of an utterance $U  \in T$. 
A good model, presented with typical text, will find it highly probable, and 
thus score a low surprisal for most utterances (tokens) in the text, and have overall low entropy. 

 Various kinds of models have been used in both natural language and in code. 
Using classical $n$-gram models of code, it has been found that entropy of
Java is around 3 to 4 bits lower than for English. 
 This is surprisingly less, suggesting that
 code is \emph{8 to 16 times more predictable than English}~\cite{Hindle2012, casalnuovo2018studying}. This gets even more noteworthy given that the \emph{vocabulary} of code, when 
 controlling for corpus size,  is typically much larger than natural language, 
 because programmers keep inventing new identifiers. Greater vocabulary \emph{could} mean  
 more word choices to spread  probability mass over---but actually doesn't! 
  Why is code so predictable?  Is it specific to Java? English? The language model? 
 Is it just because code in general has a much  simpler syntax than \nlm? Or is it somehow the result of the cognitive load of reading and writing code? 
 
 A recent paper by Casalnuovo \etal~\cite{casalnuovo2018studying} presents a detailed comparison of predictability in
 code and \nlm. They find that the greater predictability of code \emph{vs} \nlm is not specific to Java and English:
a \emph{circa} 2-5 bit (code-\nlm) difference 
persists, across  different programming languages (Java, C, Clojure, Ruby, Haskell),  different natural languages (German, Spanish, English) \emph{and} different language models ($n$gram, Cache-based, and LSTM~\cite{hochreiter1997long, sundermeyer2012lstm}). 
This persistent, robust difference  between programming and natural languages
suggests that something, perhaps the simpler syntax of code, is the cause. 
To deal with the question of syntax,  they removed keywords, operators and delimiters from code, and analogous syntactic
markers (prepositions, determiners, conjunctions, pronouns \emph{etc})\footnote{In linguistics, these are called \emph{closed category} words,
since new words in this category are very very rarely coined.} from \nlm. After the concomitant removal of these syntax markers, in code they are left merely with identifiers, and in English with just nouns, verbs, adjectives and adverbs\footnote{This concomitant removal
of syntax markers was done differently in~\cite{rahman2019natural}, which explains the different findings.}. They find that after the removal of syntax markers,  the difference between programming languages  and english {\underline {\bf \emph{increased}}} to between 4-8 bits depending on the language model used. Finally, they incorporated full parse tree structures into code (using ASTs) and English (using the Penn Tree Bank~\cite{Marcus:1993}) and found that code
still remains more predictable than English. 

Why is this?  It's certainly not because code lacks expressive power! 
Alternative forms abound even for something very simple. 
Consider the wealth of equivalent alternatives for even the trivial iteration trope:  {\small\tt for (i = 0; i < n; i = i+1) \{ $\ldots$ \}}.  One could pick a name other than {\tt i} or {\tt n}; flip the
conditional; use a different incrementing form; start and end differently, \emph{all without changing the meaning}. 
Indeed, a literal infinity of equivalent forms are possible! Still, we persist with such tropes. Why? 
To further clarify, we draw upon a formulation~\cite{barrDualTalk, allamanis2017survey}\footnote{Talk and long paper; \S 3 is most pertinent in the paper.} that software is \emph{bimodal}: it works on a human-machine channel,
and a human-human channel. It is argued that programmers  must write code with full
awareness of \emph{two modes} of eventual use: first, code has formal \emph{operational semantics}, and executes on a
machine; second, code is maintained by other programmers, who \underline{must} understand it; and thus code \emph{per se}
forms a vital communication channel between the developer who writes it and the maintainer who cares for it. 

This \emph{bimodality} argument implies that two  \emph{distinct} channels exist, and suggests
an origin for the high repetitiveness of code. First, suppose that the human-human channel 
simply re-used the formal operational semantics channel. Consider a code reader \emph{Rick}, 
who is examining a piece of code $C$, written in language ${\mathcal L}$  by developer \emph{Doris}. \emph{Rick} desires
to know its computational meaning $M$. Suppose $Rick$ could directly, by himself, mentally calculate the
meaning of $C$ via the operational semantics of $L$ quickly and easily.  
If readers can \emph{always} and \emph{efficiently} do this,
there is certainly no constraint on developer \emph{Doris}. She is free to
choose from any of the choices $C_i$ that implement $M$, since
her readers behave like infallible machines, and will reliably and efficiently find the meaning $M$, even
if she chooses bizarre (but correct) ways to implement $M$.  However,
readers are not machines, and the human-human channel works differently. 
Prior work on natural language in Psycholinguistics~\cite{rayner1996effects,wells2009experience} indicates a robust 
association between greater statistical  predictability (surprisal) and ease of production \& comprehension. 
Speakers are more likely to choose more predictable utterances, even among meaning-equivalent options~\cite{hudson2005regularizing}. These choices are likely driven in part both by \emph{audience design}~\cite{jaeger2007speakers}  (\emph{viz.,} choose utterances for ease of comprehension)
and by ease of production~\cite{bock1987effect, ferreira2000effect}. 

On the human-human channel, we might expect code to behave like natural language.  
Given a meaning, some implementation forms may be more
familiar, such as the {\small\tt for} loop above, and will be more easily recognized. 
Indeed, an experienced developer will know this, and would prefer to use familiar forms whenever possible, both
for her own and her reader's convenience.

We can formally state this thus. Assume $C_1, C_2, \ldots C_n$ are viable implementation choices for a given computation $M$. 
Although these are all semantically equivalent, for human convenience, there would be a tendency to prefer one over the
others. If  $p$  refers to probability of occurrence in a corpus, we should observe the following:
\[ \exists ~C_i, ~i \in \{1 \ldots n \}:  ~~p(C_i \mid M) \gg p(C_j \mid M)~, j \in \{1 \ldots n \}, ~j \neq i \]
We are overstating things here: it's possible that a \emph{few of the possible $C_i$} are more preferred, \emph{viz.,}
the probability mass is not uniformly spread over all the choices. Another way to state this would be that the
\emph{entropy} of this conditional distribution is less than the possible maximum which would be obtained in the
case of a uniform distribution among implementation choices. However, it is difficult to know in general the number of implementation choices. 
Given the intractability of
computing such a maximum, we formulate the central question a bit more informally: 
\begin{center}
\noindent\fbox{%
    \parbox{0.8\columnwidth}{%
For a given computation, do developers prefer some implementations over others, and to what degree? 
    }%
}
\end{center} 

We now describe our experimental decisions and specifics of our approach. 
First, in order to model the developer ``preference" above, we use \emph{language models}. 
A modern language model, well-trained
over a large, diverse corpus, can reliably\footnote{The low cross-entropy  that modern models provide
over {\underline {\emph unseen}} corpora is evidence of their power.}
capture the frequency of occurrence of textual elements 
in a corpus, thus capturing the preferences of programmers who created that corpus.

Second, we use meaning-preserving transforms to model a range of possible implementations $C_i$ for a meaning $M$. 
While many are possible, we focus on 3 types of transforms: expression rewriting, nonessential parentheses adding and removing, and variable name shuffling (for details see Section \ref{sec:semTrans}).
These transforms are performed on code fragments from a corpus ``unseen'' by  the language model, which is then used to score 
the surprisal of the original and the transformed versions. These transforms generally perform changes of a scope confined enough to be reasonably captured by our 6-gram language models
and the additional LSTM language models we use to validate effects seen on the ngram transformations.

Our first RQ investigates whether developer preferences for restricted forms of expression are observable in Java:

\RQ{rq:java}{Using a language model trained on Java code, if we perform meaning preserving transformations on unseen Java code, to what
degree does the model find the transformed code more improbable (higher surprisal)?}

Next, to ensure that this is not simply an effect of the choice of programming language, we also choose a secondary language that is  different
from Java, Python, and ask:

\RQ{rq:python}{Is the Language Model's preference for the original code also observable in Python?}

Beyond simple ngram models, we would also like to explore how local style effects the consistency of choice.  Prior research has shown this
to be a strongly distinguishing factor of source code over natural language~\cite{Tu2014, hellendoorn2017deep}, so we theorize that cache models
would prefer the original code even more strongly.  
Additionally, we would like to consider if these preference patterns are retained in the underlying structure of the code - i.e. when identifiers and literals
are abstracted.
Thus we ask:

\RQ{rq:models}{Do cache based models that incorporate local style discriminate the original code more strongly?  Is the preference for the original code
retained even when abstracting identifiers and literals?}

We expect that some transformations will disrupt the code less or even make it more probable to our models.
In particular, we would like to see how the original surprisal of the code relates to the effect of the transformation.
We would expect highly improbable code to have greater potential to become more ``typical" after transformation.
Such code will likely be associated with less restrictions on developer choice. Thus, we ask:

\RQ{rq:properties}{How does the original ``surprisal"  of the original code relate to the effect of the transformation?  
Do high-surprisal and low-surprisal code behave differently?}

Finally, to compare language model judgements with human preference
and validate model surprisal as a measure of preference in the context of code, we run a human subject study and ask:

\RQ{rq:human}{Do the preferences of Java programmers align with language model surprisal?  How do do these preferences
vary by transformation?}

\underline{{\bf In  summary:}} why is code so repetitive? Prior research clearly indicates it's not just syntax. We hypothesize that coding behavior, because of the 
the human-human channel, is susceptible to some of the same production and/or comprehension pressures as natural language. 
Thus, we expect that, for convenience, developers
strongly prefer certain forms of expression, despite the great variety of options provided by programming languages.
Moreover, these forms are \emph{predictable}, and can be captured via language models.

\section{Methodology}
\label{sec:method} 
%


We adopt a \emph{triangulation} approach, combining a 
 \emph{natural experiment} on a large code corpus with a \emph{human subject} study.  
A large code corpus embodies numerous choices made by programmers, and thus is a representative sample of these
choices. Within this corpus, we can examine the occurrence frequency of different implementation choices
$C_i$ of the same computation $M$, and determine if some choices dominate. 
For triangulation, (Section \ref{sec:human_subject_method}) we use human subjects; 
we ask them to preferentially select from two alternative implementations $C_1, C_2$ with the same meaning $M$, chosen such
that the language model assigns quite different surprisal scores to $C_1$  and $C_2$. We test
if the surprisal scores predict human preference. 

\subsection{Code Corpora}
\label{sec:data}

Our experimental dataset is chosen to help  control for potential confounds, while
also affording enough opportunities for transformations. 
Since our main focus is Java, (see RQ \ref{rq:java}) we use a larger
Java corpus and 
replicate with a smaller corpus in Python (RQ \ref{rq:python}). 

We cloned the top 1000 most starred projects on Github for Python \& Java. 
We use a subsample of these projects due to computational constraints; 
we select the 30 projects from Java \& Python with the highest count of possible transformations.
These projects are then randomly divided by project into a 70-30 training/test split. 
In Python, due to the lack of typing and limitations of the \emph{Abstract Syntax Tree (AST)}, we replicate only one type of transformation, swaps over relational operators.
These limitations also required us to normalize the original Python files with \emph{astor}\footnote{https://pypi.org/project/astor/}, which can slightly change parentheses.
As we do not perform parentheses transformations for Python, this should only minimally impact results.


Duplication can be a potentially confounding effect in training and testing code with language models~\cite{allamanisDup2018}. 
Since our focus is on programmer preferences for certain coding forms, it would be inappropriate to remove all clones. 
Still, to avoid large-scale duplicated code, we do a lightweight removal of fully duplicated files, with additional filtering during testing for stability. 
This lightweight process compares the name of every file and its parent directory (e.g. main/ExampleFile.java), keeps the first one seen of an equivalent set,
and removes the others from our training/test data.


\begin{table}[ht]
\caption{Summary of Java and Python datasets.}
\label{tab:corporaTable}
\begin{tabular}{| c | c | c | c | }
\hline
Language & Files  & Unique Files  & Training Tokens\\ \hline 
Java     & 204489 & 184093       & $\sim$118.5 M\\ \hline
Python   & 27315  & 23105        & $\sim$18.2 M  \\ \hline
\end{tabular}
\end{table}

Table \ref{tab:corporaTable} shows the file-counts and approximate token-counts in the training set for each corpora.
Duplication filtering removes 6.1\% and 10.7\% of files in Java and python respectively.
Despite sampling the same number of projects, the Java corpus is much larger, but as Java is our main focus, and Python is simply to see if the results replicate across languages, this is arguably adequate.

Our test data  was chosen to be {\underline {\emph {\bf distinct}}} from the training data. 
In addition, we removed lines commonly associated with generated code, coming from \emph{equals} and \emph{hashCode} 
functions\footnote{We dropped lines with the string's 'hashCode' or 'other' which we observed manually to be contributing to this repetitiveness. In the case of our identifier shuffling transformations which operate at the method rather than expression level (see section \ref{sec:semTrans}) we instead removed all equals and hashCode methods.}.
These lines are generated by IDEs, and arguably do not accurately represent 
\emph{human written choices} (or at least, human style choices so codified that they have been automated.).
We also remove from the test data identical lines of code appearing more than 100 times\footnote{A threshold of 10 gave similar results, suggesting robustness.}, as these may also be at risk of copy-pasting.  
We believe that it would \emph{not} be correct to simply filter out all duplicated expressions, as it is perfectly valid for developers to rewrite the same code.
Since our study is largely at the expression level, it's difficult to precisely find \& account for copy-pasting; 
we argue our approach gives a reasonable middle-ground between removing the extreme 
cases while still retaining most of the natural repetition of code. 
Finally, we note that we \underline{\emph{did not}} remove repeated or generated code \emph{fragments} from 
the \emph{training} data to properly reflect the code that programmers would read (and learn preferences from). 
Our test set pruning was to avoid overly weighting repeated and generated code, 
and emphasize more the individual, independent choices made when \emph{writing} code. 


\subsection{Language Models}
\label{sec:lm}


We estimate a language model $LM$ over a large corpus, and then use the  $P_{LM} (C)$
from the language model as an indication of developer preference. 
Specifically, we use $surprisal$ of $LM$ with respect to a fragment $C$, (which is precisely $-log(P_{LM}(C))$): 
lower surprisal indicates higher developer preference. This use of surprisal is not unprecedented; 
abundant experimental evidence from psycholinguistics indicates surprisal is strongly associated with
cognitive effort in language comprehension~\cite{hale2001probabilistic, levy2008expectation, demberg2008data, frank2013uncertainty, levy2013memory}.

We use 4 ngram language model variants to capture various aspects of possible developer preference. 
First, we use a basic 6-gram model with Jelinek-Mercer smoothing, using the best order and smoothing recommended from past research~\cite{hellendoorn2017deep}, denoted as the \emph{global} model. 
To answer the two parts of RQ \ref{rq:models}, we first use an \emph{ngram-cache} (henceforth abbreviated as \emph{cache} model), as originally described by Tu \etal~\cite{Tu2014} to capture local patterns.
Then, we build an alternate training and testing corpus where we use the Pygments\footnote{\url{http://pygments.org/}} syntax highlighter to replace all identifiers and types with generic token types,
and literals with a simplified type\footnote{For example, we keep 1,2,3 and replace higher numbers with labels like <int> and <float>. For strings, we keep the empty string, single character strings, and replace everything else with <str>.}.
These models are implemented in the SLP-Core framework by Hellendoorn \etal~\cite{hellendoorn2017deep}\footnote{https://github.com/SLP-team/SLP-Core}. 
To assess preference, we 
compare the average surprisal of tokens
that appear \emph{only in both the original and the transformed version of the expression}.  
The tokens not involved in the changed expression are not considered. 
Finally, we validate the robustness of our ngram results with a 1 Layer LSTM implemented in Tensorflow, trained with 10 epochs and 0.5 dropout,
for the corpus results of the 4 transformations used in the human study\footnote{We only considered these 4 as training and testing in context is computationally expensive, and it is just to validate the transformations we focused on in both studies.}.

\subsection{Meaning Preserving Transformations}
\label{sec:semTrans}

We choose not to use existing transformation tools, as they are either not meaning preserving (e.g. mutation testing~\cite{madeyski2010judy, just2014}),
or operate at the wrong scale of code object, such as compiler optimizations.
We use source level transformations that both are meaning-preserving 
\emph{and} small enough in scope to be captured by language models. 
Our focus is primarily on transformations of source code \emph{expressions}, which we implement via the Java and Python \emph{AST}.
For Java, we use the \emph{AST} Parser from the Eclipse Java development tools (JDT)\footnote{https://www.eclipse.org/eclipse/} and the \emph{ast}
module from Python3.7\footnote{https://docs.Python.org/3/library/ast.html}.

\begin{table}[ht]
\caption{Pseudocode examples for the transformations.}
\label{tab:transSummary}
\begin{tabular}{|c|c|c|l|l|}
\hline
\multirow{4}{*}{Swap}  & \multirow{2}{*}{Arithmetic} & *                                                          & a * b            & b * a               \\ \cline{3-5} 
                                    &                             & +                                                          & a + b            & b + a               \\ \cline{2-5} 
                                    & \multirow{2}{*}{Relational}    & ==, !=                                                     & a != b           & b != a              \\ \cline{3-5} 
                                    &                             & \textless{}, \textless{}=, \textgreater{}, \textgreater{}= & a \textless{}= b & b \textgreater{}= a \\ \hline
\multirow{2}{*}{Paren.} & \multicolumn{2}{l|}{Adding}                                                              & a + b * c        & a + (b * c)         \\ \cline{2-5} 
                                    & \multicolumn{2}{l|}{Removing}                                                            & a + (b * c)      & a + b * c           \\ \hline
\multirow{4}{*}{Rename}  & \multicolumn{2}{l|}{\multirow{2}{*}{Within Variable Types}}                              & int a            & int b               \\
                                    & \multicolumn{2}{l|}{}                                                                    & int b            & int a               \\ \cline{2-5} 
                                    & \multicolumn{2}{l|}{\multirow{2}{*}{Between Variable Types}}                             & int a            & int b               \\
                                    & \multicolumn{2}{l|}{}                                                                    & float b          & float a             \\ \hline
\end{tabular}
\end{table}

We implement 12 different kinds of transformations, summarized in Table \ref{tab:transSummary}, grouped roughly into 3 categories: 1) swapping transformations, 2) parenthesis transformations, and
3) renaming transformations.
There are 6 non overlapping sub-groups: arithmetic and relational swaps, parenthesis adding and removing, and shuffling identifiers within and between types.

\noindent{\bf Swapping transformations:} We have 8 kinds of transformations involving swapping and inverting operators, divided into 2 subcategories.  
The first subcategory swaps arithmetic operands around the commutative operators of \emph{+} and \emph{*}.
We swapped \emph{very} conservatively.  
We limit the types of the variables and literals in the expression to doubles, floats, ints, and longs.
Infix expressions with more than two operands are only transformed if the data type of the operands are int or long to avoid accuracy errors due to floating point precision limitation.
We also exclude expressions that contain function calls, since these could have side-effects that alter the other variables evaluated in the expression.

The second subcategory of operator swapping involves the 6 logical operators, \emph{==, !=, <, <=, >, >=}.
We flip the subexpressions that make up the operands for each of these, either retaining the operator if it's symmetric \emph{!=, ==},
or inverting it if it's asymmetric (e.g. \emph{>} becomes \emph{<}).
While we do not limit the types in these expressions as they are commutative and don't risk floating point issues, 
expressions with function calls are excluded to avoid side effects.

\noindent{\bf Parenthesis transformations:} The next category involves 
manipulation of extraneous parentheses in source code.
Programming languages have well-defined operator precedence, 
but programers can still (and often do) freely \emph{choose}
to include extraneous parenthesis for readability. 
For instance, in cases where less common operators are used (such as bit shifts), the parentheses may make comprehension
easier, leading to a preferred style.

Therefore, we can transform expressions by adding or removing extraneous parentheses from expressions.
The adding parentheses transformation relies on the tree structure of the AST to insert parentheses while preserving the correctness in the order of operations. 
Parentheses are not added to expressions whose parent is a parenthesized expression to avoid creating double parentheses. 
Parentheses are also never added around the entire expression.

For parentheses removal, we select each parenthesized expression.
Then, each of these are passed to the {\small\tt Necessary}\-{\small\tt Parentheses}\-{\small\tt Checker}
 from the Eclipse JDT Language Server\footnote{See \url{https://github.com/eclipse/eclipse.jdt.ls}} 
 to check if removing them would violate the order of
operations.
Any subexpressions that pass this check are then considered candidates for removal.
This method is used by the same algorithm supporting the ``Clean Up'' feature within the Eclipse IDE\footnote{See \url{https://bugs.eclipse.org/bugs/show_bug.cgi?id=405096}}.

\noindent{\bf Variable shuffling transformations:} Finally, we consider transformations that shuffle the names of identifiers.
To avoid changing meaning, we swap only within a method, using the key bindings of the \emph{AST} to maintain scoping rules.
If a variable name is used for a declaration more than once in a function (e.g. multiple loops using $i$ as a variable for iteration), it is excluded to avoid 
assigning two variables the same name within the same scope.
Methods containing lambda expressions are also ignored because their variable bindings are not available in the \emph{AST}.

We separately consider renaming both \emph{within types} and \emph{between types}.
As an example, consider a function with two \emph{int} and two \emph{String} variables.
In the \emph{within types} case, we only consider replacing one integer's name with the other, and the same for the Strings.
In the \emph{between types} case, all four variable names can be assigned to any of the integers or strings other than their original variable.
We expect that names given to the same types will be used more similarly than names given to different types.
Thus, we would expect \emph{between types} transformations result in code relatively more improbable 
than those produced by \emph{within types} transforms. 

\subsubsection{Transformation Selection}
\label{sec:trans_limits}
As expressions grow in size, the number of possible transformations grows exponentially.
Generating all these transformations are neither feasible or desirable, so we select a random subsample.
For the operand swapping and parentheses modification cases, we randomly sample up to $n$ transformations, where n is the number of possible locations to transform in the expression.
For variable renaming we consider only functions with up to 10 local variables that can be shuffled.


\subsection{Human Subject Study}
\label{sec:human_subject_method}


The corpus study can tell us if some forms are preferentially used
in a large corpus. We triangulate with a human subject study, 
checking if human preferences
over different implementations of the same computation align with different language model surprisal scores.
While lower surprisal has been linked to 
easier processing in natural language~\cite{hale2001probabilistic, levy2008expectation, demberg2008data, frank2013uncertainty, levy2013memory},  
we don't know of similar findings for code, despite the metric being used as a stand-in for preference or comprehensibility in guiding many
tools (e.g.~\cite{hellendoorn2017deep, liu2017stochastic, Allamanis2014LNC}). 


We use Amazon's Mechanical Turk (\emph{AMT})\footnote{https://www.mturk.com/} 
for this investigation into the alignment of surprisal with human-subject preference. 
\emph{AMT} has been used in several other programming studies to recruit subjects~\cite{Prana2019, Chen2019RIP, Alqaimi2019AGD}.
Since anyone can sign up with AMT, we selectively filter out a sample that can reasonably represent Java programmers. 
First, we follow recommended guidelines~\cite{TurkGuidelines} for avoiding bots and poorly qualified workers; 
we require a 99\% HIT acceptance rate, 1000 or more completed HITs, 
and restrict workers to those in the US and Canada.
We also used Unique Turker\footnote{https://uniqueturker.myleott.com/} along
with \emph{AMT}'s own internal reporting to remove any repeat users.
Secondly, we deploy a short qualification test which requires subjects to read some Java code and answer 3 comprehension questions; all 3 must be correctly answered. 
We tuned our comprehension questions with 3 pilot surveys.
MTurkers were paid at minimum wage rate (\$12/hour) for tasks they completed; the qualification test was estimated to take 5 minutes, and the main survey 20 minutes.


\noindent Our survey asked forced-choice preferences of this form: 
\begin{tabbing}
{\footnotesize\em Please select which of the two following code segments you prefer: }\\
\hspace{0.5in}\={\footnotesize\tt outPacket = new byte[10 + length];}\\

\>{\footnotesize\tt outPacket = new byte[length + 10];}\\
\end{tabbing}

\noindent 
The alternative segments were selected from the relational \& arithmetic swaps, and parenthesis adding \& removing. 
Using the \emph{global ngram model}, after some filtering\footnote{Code with hashing and bit shift keywords, lines over 80 characters.} we selected the top 20 single line transformations
that most \emph{increased} and \emph{decreased} the average {\bf line level} surprisal over shared tokens, 
replacing some from the top 40 when cases were too similar, or the transformation obviously disrupted symmetry\footnote{For example, only adding one parenthesis to a == b || b == c}.
We use line instead of expression level averages from the corpus study because the subjects judged the entire lines.
From these 160 pairs, we presented 80 to each user randomizing both the questions and order of the choices. 

To measure subject attention in a 20 minute survey, 
we included an unidentified attention check, which was a question like the others, except more obvious and incontrovertible\footnote{We asked if {\tt "for(int i = 0; i \textless~length; i++) \{" }  was preferable to {\tt "for(int i = 0; length \textgreater~i; i++) \{" }.}. 
We do not exclude those failing the attention check (it was only one question of many), instead using it as a measure for how attentive the subjects were overall.
As long as failing the check is not common, we can be confident of reasonably attentive subjects.

\subsection{Modeling}
\label{sec:meth_model}

To compare surprisal before and after the transformation, we use paired non-parametric Wilcox tests~\cite{hollander1999nonparametric}
and associated 95\% confidence intervals, which measure the expected difference in medians between the original and transformed code.
We also \emph{widen} the intervals using the conservative family-wise Bonferroni~\cite{weisstein2004bonferroni}
adjustment, to account for the tests on each model and transformation.

To answer RQ \ref{rq:properties}, we turn to regression modeling.
Recall that we theorize that expressions that are \emph{more improbable} to the language models should be more amenable to becoming more
probable after transformation, whereas \emph{low surprisal} would be associated with stronger norms and thus more harmful transformations.
We measure this effect with ordinary linear regression; we use controls for the the size of the line, the type of \emph{AST} node
that is the parent of the expression, and a summary of the operators involved in the expression\footnote{In the case of multiple operators, we selected the most common one from the training projects to represent the expression.}.
Our regressions are limited to single transformations for ease of interpretation, and filtered out rare parent and child
types ($<$ 100)\footnote{Limitations in the python transforms prevent an accurate count of the number of transformations, and so the first filter was not applied.  We examined the comparable Java models without this filter as well but found little difference in the coefficients.}.
We identified influential outliers using Cook's method~\cite{cook1982residuals} and removed those with values greater than $4/n$.
We examined residual diagnostic plots for violations of model assumptions, and made sure multicollinearity was not a issue by checking that variance inflation (VIF) scores were $<$ 5~\cite{cohen2003applied}.


For our human subject study, we use a mixed effects logistic regression. 
As the complexity of the random effects structure of our model caused the frequentist estimate to not converge, we
estimate the model via Bayesian regression through the R package brms~\cite{brms2017}.
Our presented model used the default priors of the package, but we validated convergence and alternative priors
using the guidelines included in the WAMBS checklist~\cite{depaoli2017improving}.
Further details on these models are omitted for space but can be found in the R Notebooks in our replication package (see intro of Sec. \ref{sec:results})).


\section{Results}
\label{sec:results}
Our results from the corpus study are in \S~\ref{sec:corpus_results} and the human subject study \S~\ref{sec:human_study}.
Due to space limits, our corpus study focuses on the swap transformations, and highlights differences in the other transformations. 
All our data, R notebooks, and results can be anonymously accessed at \url{https://doi.org/10.5281/zenodo.2573389}.

\subsection{Corpus Study}
\label{sec:corpus_results}

\begin{table*}[ht]
\caption{Two sided paired Wilcox signed-rank tests and 95\% confidence intervals of surprisal difference \emph{original source minus transformed source}.
A 1 bit negative difference indicates the original code is twice as probable as the transformed code. 
 Intervals are Bonferroni corrected. $\circ$ indicates $p > .05$, otherwise $p \ll .001$}
\begin{tabular}{|l|c|c|c|c|c|}
\hline
                                 & Global                    & Cache              & Global Abstracted        & Cache Abstracted         & LSTM               \\ \hline
Arithmetic Swap                  & $-0.7966, -0.7049$        & $-2.9683, -2.8249$ & $-0.5575, -0.5056$ & $-1.4682, -1.3902$ & $-0.751, -0.6271$  \\ \hline
Relational Swap (Java)              & $-1.2168, -1.1683$        & $-1.9637, -1.9053$ & $-1.7201, -1.701$  & $-1.9539, -1.9361$ & $-1.0059, -0.9517$ \\ \hline
Relational Swap (Python)            & $-1.7639, -1.6841$        & $-1.4611, -1.3783$ & $-2.481, -2.4066$  & $-1.5444, -1.4744$ & --                 \\ \hline
Add Parentheses                  & $-0.3342, -0.3202$        & $-0.5769, -0.5434$ & $-0.235, -0.2238$  & $-0.3782, -0.3641$ & $-0.543, -0.4922$  \\ \hline
Remove Parentheses               & $-0.0103, 0.0344^{\circ}$ & $-0.4388, -0.361$  & $-0.2004, -0.1497$ & $-0.4229, -0.3629$ & $-0.5552, -0.4254$ \\ \hline
Variable Shuffle (Within Types)  & $-0.8047, -0.6761$        & $-0.9738, -0.8411$ & --                 & --                 & --                 \\ \hline
Variable Shuffle (Between Types) & $-1.731, -1.6121$         & $-2.3268, -2.2314$ & --                 & --                 & --                 \\ \hline
\end{tabular}
\label{tab:CompConfInt}
\end{table*}

\subsubsection{Swapping Expressions}
\label{sec:swap_results}

\begin{figure}[ht]
\includegraphics[width=.45\textwidth]{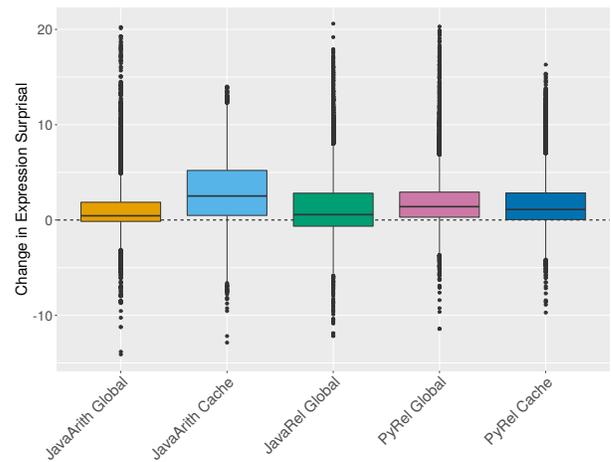}
\caption{Average surprisal change for swaps: Java arithmetic and relational, and Python relational.  Java relational cache is omitted as it is consistent with arithmetic.  Positive values indicate the transformation is less predictable.}
\label{fig:ComparisonsBoxSwap}
\end{figure}

We have 20,829 instances of transformable arithmetic swaps in our Java data, 
and 133,845/32,219 instances for 
relational swaps in Java and Python respectively.
Figure \ref{fig:ComparisonsBoxSwap} shows the difference in ngram surprisal (transformed - original) for
all of the concrete swap transformations, except for the cache for the Java relational swaps, which is similar in effect to the Java arithmetic swaps.
Rows 1,2, and 3 in Table \ref{tab:CompConfInt} show the associated Wilcox tests and confidence intervals around the median for 
\emph{all} model variants.

In general, the data supports our theory that the original code would be preferred by the language model (LM)
of the swaps to varying degrees.
For the Java arithmetic swaps, the global LM finds the original code 1.68 times more probable
(0.75 bits of surprisal less) than the transformed version, 
and the cache model finds the original 8 times more probable (3 bits less)
than the transformed version\footnote{The increase can be measured as $2^{diff}$ e.g. $2^{0.75}=1.68$.}.
The cache LM does discriminate better
across all the Java transformations, both with concrete and abstracted identifiers;  
but \emph{not}, however, in Python, perhaps due to a global ``Pythonic" culture that transcends project boundaries. 
The LSTM models we ran to validate the robustness of the ngram models on Java swaps also indicate this effect, showing
preference for the original code.  
Like the global ngram models, the effect is stronger for relational swaps.

Now, to answer RQ \ref{rq:properties}, we will describe one regression model in depth: the one for the global arithmetic swaps.
We model \emph{Surprisal Change~\textasciitilde~Original Surprisal + log(NumTokens) + ParentOperator + Operator}, 
the change in surprisal as predicted by the original surprisal, controlled by both the log size of the expression, and the
parent and operator types of the expression.
For every bit increase\footnote{Meaning the expression is twice as difficult for the model to predict.} in the original expression, the change decreases by .279 bits.
This effect is quite strong, explaining nearly 35\% of the variance in the difference.
We can conclude that less predictable expressions 
exhibit less strong norms, and the effect of a transformation is more variable.  
This negative correlation between original surprisal and the change also holds in the regressions for all the other language models on this transformation.
Among the controls, longer expressions are also more likely to be amenable to transformations, and while most parent nodes in the \emph{AST}
are similar to the '==' baseline, return statements and array accesses tend to have less strict style.
Finally, swaps that occur on a '+' instead of a '*' are 55\% less probable to the language model - likely attributable to addition being much more common than multiplication.

This effect of higher original surprisal leading to greater opportunity to make code more predictable persists across all the regressions for the Java relational swaps, 
and for the Python swap using the \emph{concrete} code.
However, in the abstracted Python code, this effects is revered, but explains only a small amount of the variance of the change (<2.5\%).
Further study is required to understand this counterintuitive behavior in the abstracted code.


So for Java swapping transformations, RQs \ref{rq:java}, \ref{rq:models}, and \ref{rq:properties} are answered affirmatively, albeit to various degrees.
The models prefer the original source code regardless of model, cache models more strongly discriminate between the original and transformed code,
and less probable expressions are associated with smaller increases, or even sometimes reductions in surprisal.
Our results for RQ \ref{rq:python} provide additional support for the overall theory, but suggest complications in the details.
The ability of the locality to discriminate may be language specific, and the relationship between the original expression surprisal and the change in surprisal in the
abstracted models (different in direction from all other results) may suggest Python norms more closely tied to identifiers.

\subsubsection{Other Transformations}
\label{sec:other_results}

\begin{table*}[ht]
\caption{Sample transformations with the largest surprisal changes for parenthesis removal with the global model.}
\begin{tabular}{|l|l|l|}
\hline
Original                                                                                                 & Transformed                                                                                            & Surprisal Change \\ \hline
double seconds = time / (1000.0);                                                                        & double seconds = time / 1000.0;                                                                        & -13.878          \\ \hline
return ((dividend + divisor) - 1) / divisor;                                                             & return (dividend + divisor - 1) / divisor;                                                             & -7.991           \\ \hline
int elementHash = (int)(element \textasciicircum (element \textgreater{}\textgreater{}\textgreater 32)); & int elementHash = (int)(element \textasciicircum element \textgreater{}\textgreater{}\textgreater 32); & 11.321           \\ \hline
c1 |= (c2 \textgreater{}\textgreater 4) \& 0x0f;                                                         & c1 |= c2 \textgreater{}\textgreater 4 \& 0x0f;                                                         & 11.053           \\ \hline
\end{tabular}
\label{tab:rmEx}
\end{table*}
For parentheses, we have 63,625 additions and 9,717 removals, with the results shown in rows 4 \& 5 in Table \ref{tab:CompConfInt}. 
The results for surprisal change, the effect of the cache, and the regression models built to answer RQ~\ref{rq:properties} are similar to those in with the swaps,
with one major exception. 
The difference between the original and transformed code in the \emph{global} model is \emph{not significant}!
We delve into this unexpected result more closely using examples
in Table \ref{tab:rmEx}.
These are fairly intuitive; the biggest improvement in predictability comes from removing parentheses unnecessary to clarify the order of operations from around a literal denominator.
In contrast, a large increase in difficulty of prediction occurs with rarely used bit shift operators---suggesting that developers may prefer parenthesis around rare 
operations to clarify order of operations.
In contrast, the LSTM models agree with our theory, although the change is smaller relative to the swaps.
Thus for we answer RQs \ref{rq:java}, \ref{rq:models}, and \ref{rq:properties} affirmatively, except for the global ngram models of parenthesis removal.
We speculate that this may be the result of less consistent style around the usage of parenthesis, 
similar to what Gopstein \etal~\cite{gopstein2017understanding, gopstein2018prevalence} found with bracket usage (see \ref{sec:rel_understanding}).
We further discuss the influence of style guidelines in \S~\ref{sec:discussion}.

Finally, we consider variable renaming transformations, 
measuring mean surprisal change across all affected expressions \emph{within the same method}.
There are 17,930 methods with shuffling within types and 
48,160 with shuffling between types\footnote{Unconstrained shuffles are possible in more methods.},
with results in rows 6 and 7 of Table~\ref{tab:CompConfInt}\footnote{As swaps operate on concrete identifiers so the abstracted models do not apply.}.
As expected, shuffling variable names within a method increases surprisal. 
Variable names matter for program comprehension, and obscuring these names is one of the most common and simple forms of program obfuscation~\cite{collberg1997taxonomy}.
Moreover, we confirm that swapping variable names across types 
is more disruptive to predictability; the difference is about twice as large.
Cache effects are still present, but diluted, possibly because the shuffle pulls its vocabulary from very similar contexts.
As with all other Java transformations, the regression models show that variable names scoring higher in surprisal have less of a surprisal increase after renaming.  
So in conclusion, the renaming shuffle transformations all answer RQs \ref{rq:java}, \ref{rq:models}, and \ref{rq:properties} as expected,
with stronger results when shuffling between rather than within types.

\begin{figure}[ht]
\includegraphics[width=.45\textwidth]{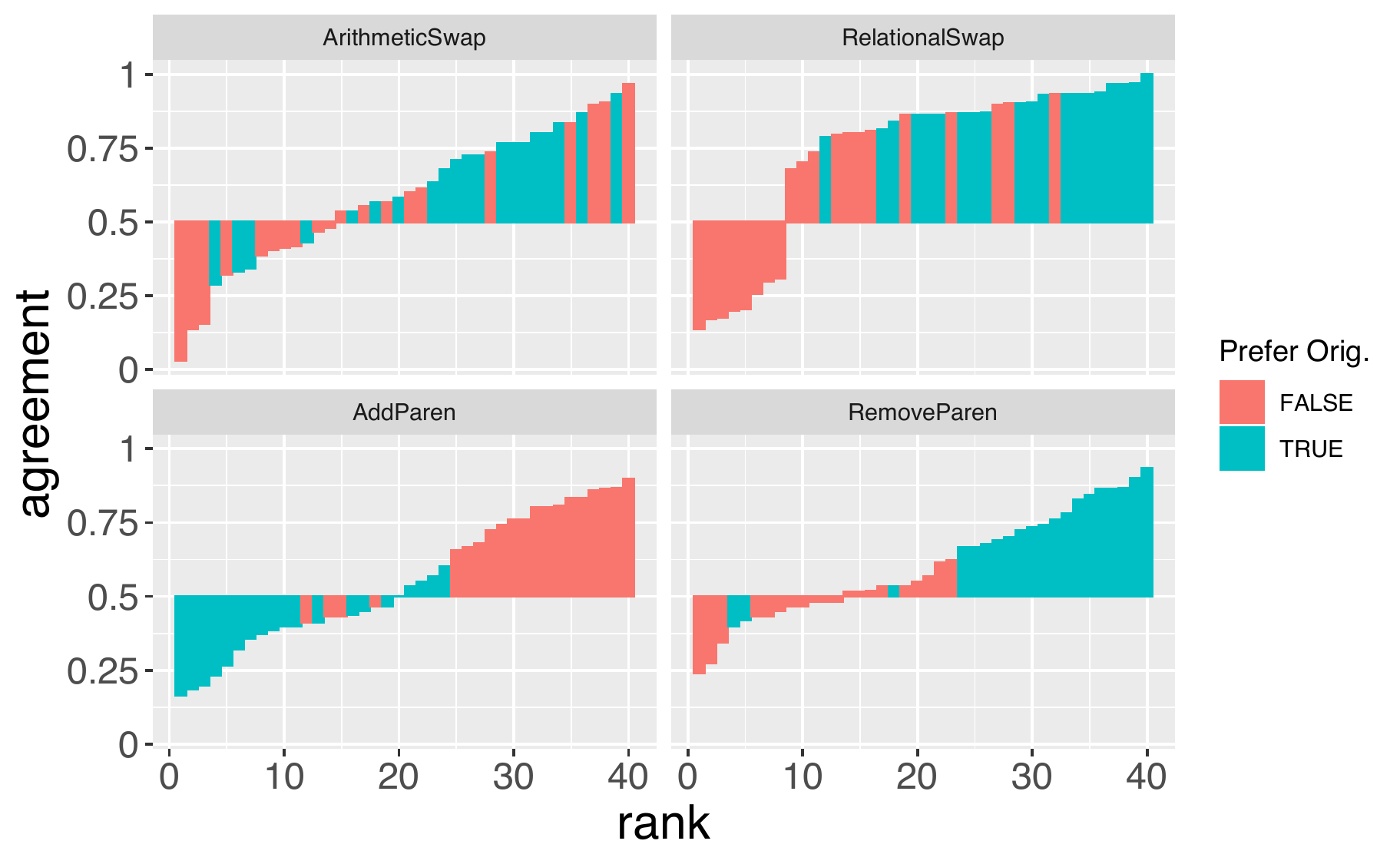}
\caption{Break down of fraction of agreement for each question by transformation, ordered from questions with the least agreement with the model to the most.  
Majority vote agreements: ArithmeticSwap (65\%), RelationalSwap (80\%), AddParen (50\%), RemoveParen (67.5\%).}
\label{fig:HSRank}
\end{figure}

\subsection{Surprisal and Human Preference}
\label{sec:human_study}


Our survey netted a total of 180 attempts across 3 batches, with 60 non-duplicate
MTurkers fully completing the survey.  Of the 60, 50 passed the attention check, though there is
little difference in overall agreement with the language model between these groups.
%
Demographically, our group had a median of 4.5 years of Java experience and 9 years of general programming experience,
and were primarily developers, students, and hobbyists who coded at least a few times a week.
All but one had some college education, and most used AMT for extra income.
%

Overall, 61.9\% (65.6\% with majority vote)\footnote{62.8 and 66.9\% for those passing the attention check.} of the time our subjects agreed with the global ngram model. 
Figure \ref{fig:HSRank} groups the results by each question, broken down by transformation type.
Each quadrant has 40 questions arranged by rank in increasing order of fraction of human agreement with the language model, 
with bars pointing downwards indicating more disagreement than agreement.
Red examples show where the language mode preferred the transformed code, and blue examples show where it preferred the original.

Humans overall agree with the language model swaps much more frequently than the parentheses changes.
Relational swaps demonstrate have the highest agreement (indicated by values above 0.5). 
All of the disagreements are cases where the raters agreed with the original code (but the language model disagreed), 
suggesting a limitation in the language model rather than disagreement among coders.
For parentheses, we also see a pattern: our group tends to prefer variants with more parentheses (indicated by the reversed red/blue patterns in AddParen and RemoveParen Figure~\ref{fig:HSRank}), regardless of the language model preference.
Moreover, the language model poorly predicts human majority vote preference for adding parenthesis - agreeing only half the time.

\begin{table}[ht]
\caption{Fixed effects for bayesian mixed effects logistic regression on our human subject study. *Parenthesis removal (RMParen) does not get an independent coefficient estimate in deviation coding, so we calculate the implied coefficient.}
\begin{tabular}{|l|c|c|c|c|}
\hline
                   & Estimate & Error & 1-95\% CI & u-95\% CI \\ \hline
Intercept          & -0.60    & 0.15  & -0.90     & -0.30     \\ \hline
LM\_Out            & 1.90     & 0.23  & 1.45      & 2.37      \\ \hline
AddParen           & -2.03    & 0.58  & -3.19     & -0.95     \\ \hline
Arithmetic         & 0.49     & 0.25  & -0.01     & 0.96      \\ \hline
Relational         & 0.15     & 0.26  & -0.39     & 0.68      \\ \hline
RMParen*           & 1.39     & --    & --        & --        \\ \hline
LM\_Out:AddParen   & -0.79    & 0.42  & -1.63     & 0.04      \\ \hline
LM\_Out:Arithmetic & -1.03    & 0.37  & -1.74     & -0.32     \\ \hline
LM\_Out:Relational & 1.64     & 0.45  & 0.77      & 2.50      \\ \hline
LM\_Out:RMParen*   & 0.18     & --    & --        & --        \\ \hline
\end{tabular}
\label{tab:HSRegression}

\end{table}

Examining the results in greater detail we present our Bayesian mixed effects logistic regression in Table \ref{tab:HSRegression}.
The model formula is: \emph{Outcome~\textasciitilde~LM\_Out * TransType + (1 + LM\_Out * TransType | ResponseId) + (1 | Question)}.
The Outcome is 1 if the human subject selected the original code, 0 for the transformation.  
The fixed effects are a binary predictor (LM\_Out), which is 1 if the \emph{language model} selected the original code, and 0 otherwise, along with the type of transformation and their interaction term.
We use the maximal random effects structure justified by the design~\cite{Barr2013} - a random intercept by question, a random intercept and slopes for surprisal, transformation type, and their intercept by subjects. 
The transformation types are deviation coded, meaning the intercept value is the grand mean over all transformations.
Thus, for the coefficients for the \emph{parenthesis removal} transformation, we subtract the 3 of other type coefficients and provide it in the table for convenience.
Finally, Bayesian estimates do not have p-values and confidence intervals in the same way as frequentist approaches.
Instead, we report the equivalent 95\% credible interval, the bounds of a probability distribution that has a 95\% probability containing the regression coefficient.
If this range is entirely above 0, it indicates a positive effect, and vice versa for a range entirely below 0.

As this is a logistic regression, the coefficients are log odds ratios.
The odds ratio of the intercept by itself, $exp(-.60) = .55$, shows that when the language model prefers the transformation, our
subjects are $1/.55 = 1.8$ times more likely to also prefer the transformed code. 
By contrast, when the language model prefers the original code, our subjects are $exp(-.60 + 1.9) = 3.7$ times more likely to also prefer the original code.
Importantly, the crucial predictor, LM\_out, has a positive effect with 0 well outside its credible interval. 
Thus, on average, not only do humans agree with the language model more often than not, they agree with the language model almost \emph{twice as strongly} in its judgements on the original code.
This effect could pose risks to tools that use surprisal to guide transformations, as the new code may not be judged as reliably by the models.
Finally, some but not all of the specific transformations also differ significantly from the grand mean in either their baseline effects or their interaction terms. 
In summary, we can say that the model confirms what was seen in Figure 2, that humans agree with the language model, except when it prefers the original code in an adding parenthesis transformation.

Therefore, our data supports RQ~\ref{rq:human}.  In most cases, the humans tend to agree with the language model. 
Adding/removing parenthesis behave differently, again suggesting that human preferences are more variable
for this transformation than for the swaps.  In particular, perhaps a more powerful language model could better capture 
preferences for adding parentheses.

%
%
%
%
%
%
%
%
%
%

\section{Discussion}
\label{sec:discussion}

Our finding that developers prefer more predictable code forms is consistent with results from 
psycholinguistics, supporting that the repetitiveness of code comes from the human-human channel. The question 
remains: why is code \emph{even more} repetitive than natural language? We propose some theories to be tested in future work.

One possible reason for the greater repetitiveness of code is that 
(unlike with code) natural language utterances are very very rarely
entirely meaning equivalent---since NL has connotative as well as propositional meaning. For instance, while ``bread and butter" is \emph{prima facie}
synonymous with ``butter and bread'', one might infer from a description 
of eating ``butter and bread'' that butter was unusually dominant in this situation.
These subtle differences of meaning may drive speakers to choose different forms in different situations. 
However,  {\small\tt i = 1+i} and  {\small\tt i =  i+1}  are semantically equivalent on the human-machine channel.  Perhaps
because developers are trained to understand the true operational semantics code, different
connotations for the above two seem unlikely to evolve, even on the human-human channel. 
Thus, it is difficult to imagine how {\small\tt i = 1+i} might carry a subtly different and useful connotation,
even on the human-human channel; and thus it is difficult find a situation (analogous to ``butter and bread'') where
a developer might consciously feel the need to use that construction. 

Another possibility is that repetitiveness is particularly beneficial in situations with increased cognitive load. Indeed, it has been proposed that children resort to repetitiveness in language learning more so than adults specifically because they have reduced cognitive capacity~\cite{hudson2005regularizing,schwab2018regularization}. Because code comprehension is challenging, repetitiveness may be extra beneficial for code compared to natural language.
Finally, repetitiveness may arise from pedagogical choices, or from coding standards.

\noindent{\bf Threats} We knowledge potential threats.  
\emph{Internal validity} threats might arise from a few sources. First, our transforms must be sound. 
We carefully reviewed the code of and hand-checked a large sample of the results to ensure correctness and diminish the possibility of error. 
Second, while we primarily used lexical ngram based models, we validated the robustness of our corpus study with LSTM models on several examples.
Moreover, for the small-scope localized transforms we use, we believe they are adequate, as such models have been used in prior work
of this nature~\cite{casalnuovo2018studying}.  

Regarding \emph{external validity}, one issue is our choice of projects. 
We have chosen a reasonable sample of projects in two widely-used languages, and our results largely hold up.  We believe it's likely that
our results will generalize to languages similar to Java and Python. 
It's possible that other languages (\emph{e.g.} Haskell) with tightly-knit, highly-skilled user-bases may behave differently. 
We have also only focused on small-scale transformations to expressions. 
It's possible that that larger transforms may have different effects, and may require different modeling techniques as mentioned above. 
Finally, coding style guides can influence how code is written. 
We searched our projects for references to style guides and found several variants.
Some projects had explicit style checks, and others were much looser\footnote{For instance, Apache Tomcat.}. 
Virtually all the guidelines were largely unrelated to our transformations, and more focused on naming \& whitespace. 
We did find a few, limited references in Java to using parentheses as needed for clarity; and just one project specified that \emph{null} values should come second, but otherwise nothing that would affect our transforms. 
\emph{Construct validity}: we measure prevalence using surprisal from language models. 
Language models are highly-refined methods for estimating occurrence frequency, and have proven value in natural language processing, so this is well-justified. 

\noindent{\bf Actionability and Future Work:} We acknowledge that our work thus far is more science than engineering, but it does have practical implications. 
While sound, meaning-preserving transforms aren't realistic for natural language, they are for code! 
We confirm that surprisal of code is strongly associated with human preference, thus providing theoretical support for a 
tool that aims to rewrite code into a meaning-preserving form preferred by programmers. 
We plan to see if better neural models, such as the Transformer~\cite{vaswani2017attention}, align with human preference more strongly.
Psycholinguistic experiments have demonstrated that natural language comprehension and production are robustly sensitive to surprisal across a wide range of measures (e.g. forced-choice preferences, reading times, production times, comprehension accuracy, neural measures, etc.)~\cite{rayner1996effects, wells2009experience,morgan2016abstract, oldfield1965response, kutas1984brain}. In future work we plan to extend the human subjects work here to test whether 
ease of code comprehension and production similarly relates to surprisal. 

\section{Related Works}
\label{sec:rel_works}

First, we briefly note the recent work by Rahman \etal claiming the greater repetitiveness of source code over English is diminished once syntactic tokens are removed~\cite{rahman2019natural}. 
They 
compared a fixed baseline of English without syntactic markers against code with and without these tokens.
However, Casalnuovo \etal performed \emph{two} pairwise English-Java corpus comparisons, with and without
syntactic (closed-category) tokens. 
This more balanced comparison reveals that  \emph{without the markers of syntax},
the gap of predictability between code and English actually \emph{increases}~\cite{casalnuovo2018studying}.

\subsection{Program Understanding}
\label{sec:rel_understanding}

%
%
%

While we draw theoretical inspiration from the ``bimodality" of software~\cite{barrDualTalk, allamanis2017survey}, 
we also note that this theory is only a recent re-formulation of a much older idea.
The idea of programs serving a dual purpose, for machine and human, is decades old~\cite{brooks1978using}.
In the study of program understanding, this connects to the ideas of \emph{top-down} and \emph{bottom-up} comprehension.
\emph{Top-down comprehension} arguably relates the human-human channel, where past experience guides a reader to seek out
expected cues called \emph{beacons}, that help her decipher the program's meaning~\cite{brooks1978using, brooks1983}.
In contrast, \emph{bottom-up comprehension} involves processing individual pieces of code, 
storing them in memory as \emph{semantic chunks} and
constructing the meaning out of these pieces~\cite{pennington1987, shneiderman1979syntactic}.
This in some ways resembles the way a machine would process code, where
understanding arises out of precise operational semantics. 

Program understanding using fMRI\footnote{Functional Magnetic Resonance Imaging.} and eye-tracking as humans read programs has seen recent focus.
A study by Seigmund \etal used fMRI to study both top-down and bottom-up comprehension
in a programming environment~\cite{siegmund2017measuring}. 
Using brain activation to measure ``neural efficiency" (which associates
lower brain activation with greater cognitive ease~\cite{neubauer2009intelligence, siegmund2017measuring}), 
they find that top-down comprehension \emph{is} more efficient than bottom-up comprehension. 
This supports the theory that the availability of highly probable ``beacons" expected
by humans facilitates code reading on approaches that rely on bottom-up construction of
semantics. 

Meanwhile, eye-tracking studies (see survey by Obaidellah \etal.~\cite{obaidellah2018survey}) also help explain 
how humans understand code and how they do so \emph{differently} from natural language.
For instance, studies show that while natural language follows a linear reading pattern (left to right in English), 
code readers jump around quite a bit,
\emph{e.g.} from a variable or function use to it's declaration~\cite{busjahn2015eye, jbara2017programmers}.  
Indeed, expert readers tend to show more non-linear eye traces than novices~\cite{jbara2017programmers}.
These techniques have also been used together: Fritz \etal used eye-tracking in combination with 
EEG-based measures of electrical brain activity, to predict the difficulty of programming tasks~\cite{fritz2014using}, and recent calls for similar studies combining eye and brain methods highlight their potential for
understanding program comprehension~\cite{peitek2018toward}. 
Fakhoury \etal used fNIRS (similar to fMRI) and eye tracking~\cite{Fakhoury2019} to relate cognitive load to lexical but not structural anti-patterns in code, though both led to worse task performance.
Finally, in natural language, surprisal is known to relate
to eye movement and comprehension~\cite{hale2001probabilistic, levy2008expectation, demberg2008data, frank2013uncertainty, levy2013memory}; 
our work suggests that lower surprisal in code will also ease reading \& comprehension, which we hope to pursue in future work. 

Finally, we note that
Gopstein \etal~\cite{gopstein2017understanding, gopstein2018prevalence}  found 
that style guides advocating for using only necessary curly braces aren't empirically well founded; sometimes
superfluous braces aid program understanding. 
This is consistent with our finding that parentheses are preferentially included to indicate
evaluation order (even when not needed) to
serve a similar role in segmenting expressions as curly braces do in control flow.
Although the predictability of source code and human understanding are different metrics, 
our results on parentheses removal suggest a similar
phenomenon - that developers use them sometimes to benefit from easier readability. 

\subsection{Generated \emph{vs.} Stored Language}
\label{sec:rel_gen_language}

As we focused on repeated patterns used to express the same meaning, 
we highlight work on natural language \emph{exemplar} theories.
These \emph{exemplars} are examples (at the word or phrase level) that are learned from usage and then generalized~\cite{bybee2003phonology,hay2006spoken}.
Importantly, some phrases are neither stored entirely in memory nor strictly generated from grammar, forming a category in between the two.
Recent work by Morgan \etal examined this on order preferences in binomial expressions, such as \emph{bread and butter} vs \emph{butter and bread}, 
to see how frequency of usage and abstract linguistic preferences\footnote{As an analogy, an abstract linguistic preference in code might be that variables go before constants, \emph{e.g.,} {\small\tt i+1} rather than {\small\tt 1+i}.} 
determined what humans prefer.
They found that both effects played roles in preference, but frequency of expression overwhelms the effects of underlying preferences and codifies a norm~\cite{morgan2016abstract}.
We use similar human subject study, but leverage code's structure to combine this with a natural experiment on code corpora.

We also briefly note a specific type of stored language, \emph{idioms}, as they have been explored in software before.
Idioms can quickly and efficiently convey a meaning for to those who know them~\cite{schmitt2004formulaic}. 
Idiomatic language has been mined from source code by Allamanis \etal, finding syntactic fragments across programs that possess the same meaning~\cite{Allamanis2014Idioms},
though they focus on extracting them over studying controlled preferences as we do.

\label{sec:rel_prog_variants}

Finally, we briefly note recent work in approaches to generating program variants in mutation testing and program obfuscation.
Mutation testing seeks to create semantically \emph{different} programs to expose deficits in test suites~\cite{papadakis2019mutation}.
One relevant work has tried using language models to find \emph{natural} mutants, finding that the mutants tended to be less natural (more improbable)
but did not have success using the metric to guide mutation selection~\cite{jimenez2018mutants}.
Additionally, obfuscation transformations generally retains meaning, though they can produce different error behavior and run much slower~\cite{collberg1997taxonomy, collberg1998manufacturing}.
Recent work~\cite{liu2017stochastic} has used naturalness to combine obfuscation operators in a way to minimize the effectiveness of deobfuscation techniques learning patterns out
of software~\cite{Raychev2015, vasilescu2017recovering}.
These approaches, however, are more focused on applications than understanding the decisions that inform source code choices.

\section{Conclusion}
\emph{Why is code so repetitive?} Previous work strongly suggests that it is not merely the restricted syntax of
programming languages; and it's most definitely not because programming languages restrict the possible 
ways to express computations. In this study, we hypothesize
that programmers prefer certain ways to write code. We model familiarity using a language
model estimated over a large corpus, and measure the "familiarity" of different ways writing code, while controlling
for the meaning. We find that \emph{``familiar"} forms are indeed more preferred by code writers, using surprisal, as scored by a language model, 
and also align with preferences of human readers via controlled experiment on Mechanical Turk. 
Finally, we draw connections between our work and the well-established 
theories from Psycholinguistics.

\bibliographystyle{ACM-Reference-Format}
\bibliography{semtrans}

\end{document}